\documentclass{article}

     \PassOptionsToPackage{numbers, compress}{natbib}

  \usepackage[preprint]{neurips_2019}

\usepackage[utf8]{inputenc} 
\usepackage[T1]{fontenc}    
\usepackage{hyperref}       
\usepackage{url}            
\usepackage{booktabs}       
\usepackage{amsfonts}       
\usepackage{nicefrac}       
\usepackage{microtype}      
\usepackage{graphicx}
\usepackage{cite}
\usepackage{multirow,setspace,verbatim,amsfonts,graphicx,amsmath,amsbsy,amssymb,epsfig,url}
\usepackage{gensymb}
\usepackage{booktabs}
\usepackage{algorithm}
\usepackage[noend]{algpseudocode}
\usepackage{amsfonts}
\usepackage{amsmath,bm}
\usepackage{amssymb}
\usepackage{tikz,graphicx}
\usepackage{mathrsfs} 
\usepackage[algo2e]{algorithm2e} 
\usepackage{array,multirow}
\usepackage{color}
\usepackage{caption}
\usepackage{subcaption}

\newtheorem{lemma}{\bf{Lemma}}
\newtheorem{proposition}{\bf{Proposition}}

\def\QED{~\rule[-1pt]{5pt}{5pt}\par\medskip}
\newenvironment{proof}{\emph{Proof.}}{\hfill\QED}

\long\def\comment#1{} 

\newcommand{\xmath}[1] {\ensuremath{#1}\xspace}
\newcommand{\blmath}[1] {\xmath{\bm{#1}}}
\newcommand{\A}{\blmath{A}}
\newcommand{\B}{\blmath{B}}

\newcommand{\E}{\blmath{E}}

\newcommand{\x}{\blmath{x}}

\newcommand{\y}{\blmath{y}}

\newcommand{\z}{\blmath{z}}
 
\newcommand{\0}{\blmath{0}}

\newcommand{\Db}{{\blmath D}}
\newcommand{\Eb}{{\blmath E}}
\newcommand{\Fb}{{\blmath F}}
\newcommand{\Gb}{{\blmath G}}

\newcommand{\Ib}{{\blmath I}}

\newcommand{\Lb}{{\blmath L}}

\newcommand{\wb}{{\blmath w}}
\newcommand{\xb}{{\blmath x}}

\newcommand{\Phib}{{\boldsymbol {\Phi}}}

\newcommand{\Xib}{{\boldsymbol {\Xi}}}
\newcommand{\Thetab}{{\boldsymbol {\Theta}}}
\newcommand{\Sigmab}{{\boldsymbol {\Sigma}}}
\newcommand{\Rd}{{\mathbb R}}

\newcommand{\mub}{{\boldsymbol{\mu}}}
\newcommand{\phib}{{\boldsymbol{\phi}}}
\newcommand{\psib}{{\boldsymbol{\psi}}}

\newcommand{\Ybc}{{\boldsymbol{\mathcal Y}}}

\newcommand{\Zbc}{{\boldsymbol{\mathcal Z}}}

\newcommand{\Xbc}{{\boldsymbol{\mathcal X}}}

\newcommand{\beq}{\begin{equation}}
\newcommand{\eeq}{\end{equation}}
\newcommand{\beqa}{\begin{eqnarray}}
\newcommand{\eeqa}{\end{eqnarray}}

\title{Boosting CNN beyond Label in Inverse Problems}

\author{%
  Eunju Cha$^\textbf{1}$  \\
    \And
   Jaeduck Jang$^\textbf{2}$\\
  \And
  Junho Lee$^\textbf{2}$\\
 \And
  Eunha Lee$^\textbf{2}$\\
  \And
   Jong Chul Ye$^\textbf{1}$\thanks{Jong Chul Ye is the corresponding author of this paper.} \\  
  \AND
  \texttt{$^\textbf{1}$ Department of Bio and Brain Engineering, KAIST, South Korea}  \\ 
     \texttt{$\{$eunju.cha, jong.ye$\}$@kaist.ac.kr} \\
  \texttt{$^\textbf{2}$ Samsung Advanced Institute of Technology} \\
}

\begin{document}

\maketitle

\begin{abstract}
Convolutional neural networks  (CNN)  have been extensively used for  inverse problems.
However,  their prediction error  for unseen test data is difficult to estimate  a priori since the neural networks are
trained using only selected data and their architecture are largely considered a blackbox.  This poses a fundamental
 challenge to neural networks for unsupervised learning or improvement
beyond the label.
In this paper, we show that the recent unsupervised learning  methods such as Noise2Noise,
 Stein’s unbiased risk estimator (SURE)-based  denoiser, and
Noise2Void  are closely related to each other in their formulation of an unbiased estimator of the prediction error,
but each of them are associated with its own limitations.
%
Based on these observations,  we provide a novel boosting estimator for the prediction error.
In particular,
by employing combinatorial convolutional  frame representation of encoder-decoder CNN
and synergistically  combining it with the batch normalization, 
we provide a  close form formulation for the unbiased estimator of the prediction error  that can be minimized for neural network training beyond the label.
Experimental results show that the resulting algorithm, what we call Noise2Boosting, provides consistent improvement in various inverse problems under both supervised and unsupervised learning setting.
\end{abstract}

\section{Introduction}

Suppose that the measurement $\xb$  is corrupted with  additive noises:
\begin{align}\label{eq:xmodel}
\x = \mub + \wb ,\quad \wb \sim \mathcal{N}(\0, \sigma^2\Ib)
\end{align}
where $\mub \in \Rd^n$ denotes the unknown mean vector, and $\sigma^2$ is the variance.
%
%
Consider a deep neural network model $\Fb(\x):=\Fb_\Thetab(\x)$ with the weight $\Thetab$, which is   trained with the data $\x$ producing
an estimate $\widehat\mub=\Fb_\Theta(\x)$. Then, our goal is to estimate the {\em prediction error}:
\begin{eqnarray}\label{eq:PE}
\mathrm{Err}(\widehat\mub) = E_{\x,\x^*}\|\x^*-\widehat\mub \|^2 =  E_{\x,\x^*}\|\x^*-\Fb(\x) \|^2
\end{eqnarray}
which quantifies how well $\widehat\mub$ can predict a
test data $\x^*$,  independently drawn from the same distribution of $\x$ \citep{efron2004estimation,tibshirani2018excess}.

The problem of estimating the prediction error is closely related to the generalizability of neural network \citep{anthony2009neural}. Moreover, this problem is tightly
linked to the classical approaches for  model order selection in statistics
literature \citep{stoica2004model}. For example, one of the most investigated statistical theories to address
this question  is so-called covariance penalties approaches such as  
Mallow’s Cp \citep{mallows1973some}, Akaike’s information criterion (AIC) \citep{akaike1974new}, Stein’s
unbiased risk estimate (SURE) \citep{donoho1995adapting}, etc. 

This paper is particularly interested in estimating the prediction error 
of encoder-decoder convolutional neural networks (E-D CNNs) such as U-Net \citep{ronneberger2015u,han2018framing,ye2019cnn}. 
E-D CNNs 
have been extensively used for various inverse problems
such as image denoising,  superresolution, medical imaging, etc \citep{ye2018deep}.
Recent theoretical results showed that, 
thanks to the ReLU nonlinearities,
the input space
is partitioned into the non-overlapping regions so that input images in each region share the same linear frame representation, but not across different partitions \citep{ye2019cnn}.
In this paper, we will show that this property can be exploited  to derive an explicit formulation for the unbiased
estimator of the prediction error. 

Our explicit formulation reveals an important link to the existing
unsupervised denoising networks.
For example, an unsupervised training scheme
called Noise2Noise (N2N) \citep{lehtinen2018noise2noise} is based on the training between multiple realizations of noisy images without clean reference data.
We show that  the loss used by Noise2Noise  is indeed an unbiased estimator of the prediction error when
 many independent noisy realizations are available for each image.
When only one noisy realizations are available, 
Stein’s unbiased risk estimator (SURE)-based  denoiser \citep{soltanayev2018training} is shown as the unbiased estimator for the prediction error.
Unfortunately, aside from the inconvenience of using MonteCarlo method to estimate the divergence term \citep{ramani2008monte},
it is often difficult to prevent the network from  learning a trivial
identity mapping.
We show that another denoising scheme, known as
Noise2Void \citep{krull2018noise2void}, can partially overcome the limitation of SURE-denoiser,
thanks to the boostrap sampling of the input data \citep{efron1994introduction}, which prevents the identity mapping from being learned.
Nonetheless, the use of subsampled data as input makes the network performance limited.

To address this problem, here we provide a novel boosting estimator for the prediction error that can be used for neural network training for both
supervised and unsupervised learning problems.
Moreover, with proper batch normalization \citep{ioffe2015batch,hoffer2018norm,cho2017riemannian,miyato2018spectral,ulyanov2016instance}, we  show that the contribution of the
divergence term for the resulting loss can be made trivial, which can significantly
simplify the neural network training.
The resulting algorithm, what we call the {\em Noise2Boosting}, has many advantages.
In contrast to Noise2Noise that requires multiple perturbed noisy images as targets,
our framework only requires one target image. 
Unlike the Noise2Void, multiple neural network output from bootstrap subsamples or random weighted input images  are adaptively combined to provide
 a final output, which can reduce the prediction error.
 The applicability of the new method is demonstrated using various inverse problems such 
 as denoising, super-resolution, accelerated MRI, electron microscopy imaging, etc, which
 clearly show that our Nose2Boosting can significantly improve the image quality beyond labels.
All  proofs for the lemma and propositions are provided in Supplementary Material.

\section{Related works}

In Noise2Noise \citep{lehtinen2018noise2noise}, the main assumption is that
a neural network may learn to output the average of all feasible explanations.
Therefore, rather than using the ground-truth reference, the authors claimed
that any noisy data from the same distribution can be used as a target
for the neural network training.
However, N2N requires multiple noisy images for the same underlying noiseless image during the training,
which may not be feasible in some acquisition scenario.
To address this problems, the authors in \citep{krull2018noise2void} proposed
so called Noise2Void (N2V) training scheme that utilizes
random masked images as  input.
The main assumption for N2V is that each pixel can be predicted from its neighbors using a neural network
  different from the identity mapping. Unfortunately, N2V has limitations
in capturing fast varying and isolated structures.
Instead, the authors in \citep{soltanayev2018training} proposed SURE-based denosing network
that employs the divergence penalty to regularize the auto-encoder loss.
Due to the difficulty of obtaining divergence penalty,  the authors in \citep{soltanayev2018training} employed
the MonteCarlo SURE \citep{ramani2008monte} to approximate the divergence.

Bootstrap aggregation (bagging) \citep{breiman1996bagging} is a classical machine learning technique which uses
bootstrap sampling and aggregation of the results to reduce the variance  to improve the accuracy of the base learner.
Boosting \citep{schapire1999brief} differs from bagging in their multiplication of  random weights to the training data rather than subsampling.
The rationale for bagging and boosting is that it may be easier to train
several simple weak leaners and combine them into a more complex learner than to
learn a single strong learner.

\section{Theory}

This section analyzes  Noise2Noise, SURE-based denoiser, and Noise2Void from the perspective
of their capability of estimating the prediction error. This leads to a novel boosting scheme.

\subsection{Existing approaches}

Recall that  the unseen test data $\x^*$ in \eqref{eq:PE}
can be represented by
\[\x^*=\mub+\wb^*,\]
where $\wb^*$ is drawn from
$\mathcal{N}(\0, \sigma^2\Ib)$ and 
independent  from $\wb$ in \eqref{eq:PE}.
This suggests an  estimator for the prediction error  in \eqref{eq:PE}:

\begin{eqnarray}\label{eq:N2N}
\widehat{\mathrm{Err}}(\x,\x^*) = \|\x^* - \Fb_\Thetab(\x)\|^2
\end{eqnarray}

We can easily see that
$E_{\x,\x^*}\left[\widehat{\mathrm{Err}}(\x,\x^*)\right] = E_{\x,\x^*}\|\x^* - \Fb_\Thetab(\x)\|^2$,
suggesting that $\widehat{\mathrm{Err}}(\x,\x^*)$ is an unbiased estimator of the prediction error in \eqref{eq:PE}.
In fact, the Noise2Noise  estimator is an extension of \eqref{eq:N2N} for the
training samples $\{\x^{(i)},\x^{*(j)}\}_{i,j=1}^{P,Q}$, resulting in the following loss:
\begin{eqnarray}
\ell_{N2N}(\Thetab) := \frac{1}{PQ}\sum_{i=1}^P\sum_{j=1}^Q \|\x^{*(j)} - \Fb_\Thetab(\x^{(i)})\|^2
\end{eqnarray}
Accordingly,  the neural network training is done by  minimizing the loss:
\[\widehat\Thetab= \arg\min_\Thetab \ell_{N2N}(\Thetab).\]
While $\ell_{N2N}(\Thetab)$ is also an unbiased estimator of the prediction error \eqref{eq:PE}, due to independent sampling along $\x$ and $\x^*$,
large number of noisy samples are required to reduce the variance of the estimator $\ell_{N2N}(\Thetab)$.

On the other hand, the authors in \citep{soltanayev2018training} employed the Stein Risk Unbiased Estimator (SURE).
The original formulation was derived for the unbiased {risk} estimator, i.e. $E\|\mub-\Fb_\Thetab(\x)\|^2$,  but can be easily
converted for the prediction error estimator. Accordingly,
the corresponding SURE for the prediction error  becomes
\begin{eqnarray}
\widehat{\mathrm{SURE}}(\x) :=  \|\x - \Fb(\x)\|^2+ 2\sigma^2 \mathrm{div} \{ \Fb(\x) \}
\end{eqnarray}
where $\mathrm{div}(\cdot)$ denotes the divergence.
Then, the SURE denoising network training is  done for the
training samples $\{\x^{(i)}\}_{i=1}^{P}$ by minimizing the following loss:
\begin{eqnarray}\label{eq:SURE_dn}
\ell_{SURE}(\Thetab):=
\frac{1}{P}\sum_{i=1}^P \|\x^{(i)} - \Fb_\Thetab(\x^{(i)})\|^2 + 2\sigma^2 \mathrm{div} \{ \Fb_\Thetab(\x^{(i)}) \}
\end{eqnarray}
Although the application of SURE for unsupervised denoising is an important advance in theory, there are several
practical limitations. First, due to the difficulty of calculating the divergence term, the authors relied
on MonteCarlo SURE \citep{ramani2008monte} which calculates the divergence term using MonteCarlo simulation.
This introduces additional hyperparameters, on which the final results critically depend.
Another important drawback of SURE denoising network is that it is difficult to
prevent the network from learning a trivial identity mapping.  More specifically, 
if $\Fb_\Thetab(\x)=\x$, the cost function in \eqref{eq:SURE_dn} becomes zero.
One way to avoid this trivial solution is to guarantee that the divergence term at the optimal network parameter
should be negative. However, given that the divergence term comes from the degree of the freedom \citep{efron2004estimation} and the amount of
excess optimism in estimating the prediction error \citep{tibshirani2018excess}, enforcing negative value may be unnatural.

Recently,  the authors in \citep{krull2018noise2void} proposed so-called Noise2Void (N2V).
In N2V, various position of the input pixels are masked with random distribution.  Let $L$ denote the random variable for the sampling
index and the corresponding neural network estimate is referred to as $ \Fb_\Thetab(\x,L)$.
Then, the corresponding prediction error is given by
\begin{eqnarray}\label{eq:PEL}
\mathrm{Err}(\widehat\mub) = E_{\x,\x^*,L}\|\x^*-\Fb_\Thetab(\x,L) \|^2
\end{eqnarray}
and the associated SURE estimator is given by
\begin{proposition}\label{prp:SURE_L}
The SURE estimate  
\begin{eqnarray}\label{eq:SURE_L}
\widehat{\mathrm{SURE}}(\x,L) :=  \|\x - \Fb_\Thetab(\x,L)\|^2+ 2\sigma^2 \mathrm{div} \{ \Fb_\Thetab(\x,L) \}
\end{eqnarray}
is an unbiased estimate for the prediction error in \eqref{eq:PEL}.
\end{proposition}

 For the given training samples  $\{\x^{(i)}\}_{i=1}^{P}$ and random mask $\{L_k\}_{k=1}^K$,
the oise2Void denoising network training could be  done by minimizing the
following loss:
\begin{eqnarray}\label{eq:N2V}
\ell_{N2V}(\Thetab):=
\frac{1}{PK}\sum_{i=1}^P\sum_{k=1}^K \|\x^{(i)} - \Fb_\Thetab(\x^{(i)}, L_k)\|^2 + 2\sigma^2 \mathrm{div} \{ \Fb_\Thetab(\x^{(i)}, L_k) \}
\end{eqnarray}
Similar to N2N estimator, N2V estimator requires sampling along $\x$ and $L$.
However, the main goal of the index subsampling  is to prevent from learning identity mapping;
therefore, the number of index sampling is usually small.

In the original N2V, the divergence term in \eqref{eq:N2V} was not used for training.
Later, we will show that with the batch normalization, the
contribution of the divergence term can be made trivial, which again explains the success of N2V. 
%
%
%

\subsection{Bagging/Boosting method}

Based on the discussion so far, here we propose a novel  boosting method,
where the input data are subsampled  by bootstrapping or multiplied with random weights,  and the final output is the aggregated value.  
More specifically,  the bagging (or boosting) provides the network output as an average of the network trained with
bootstrap subsampled input (or random weighted input):
\begin{eqnarray}\label{eq:bagging}
\widehat\mub(\x) &=& E_L\{\Fb(\x,L)\} 
\end{eqnarray}
Since the dependency on $L$ is integrated out in \eqref{eq:bagging} during the aggregation,
the corresponding prediction error takes the same form as in \eqref{eq:PE}.
Then, we have the following key result:

\begin{proposition}\label{prp:PE_bagging}
\[ E_{\x,\x^*,L}\|\x^* - \Fb(\x,L) \|^2 \geq  E_{\x,\x^*} \| \x^* - E_{L}\{ \Fb(\x,L) \} \|^2 \]
\end{proposition}

Proposition~\ref{prp:PE_bagging} show that the prediction error of boosting can be better than that of the N2V.
Moreover, the gap increases more when the $\Fb(\x,L)$ provides diverse output for each realization of the index $L$ \citep{breiman1996bagging}.
In fact, due to the combinatorial nature of ReLU,  it was shown in \citep{ye2019cnn} that the input space results in non-overlapping partitions with different
 linear representations. Therefore, by changing the subsampling pattern $L$, the distinct representation
may be selected, which can make the corresponding neural network output $\Fb(\x,L)$ diverse.

Finally,  it is straightforward to see that the  SURE estimator for the associated prediction error is given by
\begin{eqnarray}
\widehat{\mathrm{SURE}}(\x) := \left\| \x - E_L\{\Fb(\x,L)\}  \right\|^2+ 
{2\sigma^2} \mathrm{div} E_L\{ \Fb(\x,L)  \}
\end{eqnarray}

Then, for the given training samples  $\{\x^{(i)}\}_{i=1}^{P}$,
the boosted  network training can be  done by minimizing the loss:
\begin{eqnarray}\label{eq:N2B}
\ell_{Boosting}(\Thetab):=
\frac{1}{P}\sum_{i=1}^P \|\x^{(i)} - E_L\{\Fb_\Thetab(\x^{(i)},L)\}\|^2 + 2\sigma^2 \mathrm{div} E_L\{ \Fb_\Thetab(\x^{(i)}, L) \}
\end{eqnarray}


\section{Noise2Boosting: A Novel Boosting Scheme}

Based on the theoretical analysis so far, we now introduce an efficient implementation of \eqref{eq:N2B}, what we call
the {\em Noise2Boosting (N2B)}.  The proposed N2B method is based on the following two simplifications.

\subsection{Divergence  simplification}

In SURE estimator,  calculation of the divergence term
is not trivial for general neural networks. 
This is why the authors in \citep{soltanayev2018training}  employed the MonteCarlo SURE.
In this section, we first show that  there exists a simple explicit form
of the divergence term for the case of E-D CNNs.
Then, the batch normalization is shown to make   the divergence term trivial.


Specifically, as shown in \citep{ye2019cnn},  the output of the ED-CNN can be represented by nonlinear basis representation:
\begin{eqnarray}\label{eq:basis}
\y := \Fb(\x) 
= \sum_{i} \langle {\blmath b}_i(\x), \x \rangle \tilde  {\blmath b}_i(\x) = \tilde \B(\x)\B(\x)^\top \x
\end{eqnarray}
where
 $ {\blmath b}_i(\x)$ and $\tilde  {\blmath b}_i(\x)$ denote the $i$-th column of the following frame basis and its dual:
\begin{eqnarray}
\B(\x)&=& \Eb^1\Sigmab^1(\x)\Eb^2 \cdots  \Sigmab^{\kappa-1}(\x)\Eb^{\kappa},~\quad \label{eq:Bc}\\
\tilde \B(\x) &=& \Db^1\tilde\Sigmab^1(\x)\Db^2 \cdots  \tilde\Sigmab^{\kappa-1}(\x)\Db^{\kappa} \label{eq:tBc}
\end{eqnarray}
where
$\Sigmab^l(\x)$ and $\tilde\Sigmab^l(\x)$ denote the diagonal matrix with 0 and 1 values that are determined by the ReLU output
in the previous convolution steps, $\Eb^l$ and $\Db^l$ refer to the encoder and decoder matrices, respectively,
 whose explicit forms can be found in  Supplementary Material.
Since the patterns of $\Sigmab^l(\x)$ and $\tilde\Sigmab^l(\x)$ depend on the input,  the expression suggests
that the input space is partitioned into multiple regions where input signals for each region share the same linear representation, but not across
different partitions.

Using this,  we can easily obtain the close-form expression for the divergence term.
\begin{lemma}\label{lem:div}
Let $\Fb(\x)$  be represented by \eqref{eq:basis}. Then, we have
\begin{eqnarray}
 \mathrm{div} \{ \Fb(\x) \} =     \sum_i  \langle {\blmath b}_i(\x), \tilde  {\blmath b}_i(\x) \rangle
\end{eqnarray}
\end{lemma}

\begin{proposition}\label{prp:Ediv}
Suppose the index set $L$ is obtained by the either  1) sampling with replacement  such that each index can be selected
with the probability of $p$, or 2)  random weighting with the mean value of $p$.  Then,
\begin{eqnarray}
 \mathrm{div} E_L\{ \Fb(\x,L) \} =    p\sum_i  \langle {\blmath b}_i(\x), \tilde  {\blmath b}_i(\x) \rangle
\end{eqnarray}
\end{proposition}


Proposition~\ref{prp:Ediv} informs that the divergence term for the boosted estimator can be simply
represented using  the nonlinear frame of the E-D CNN.
This leads to a
simple approximation of divergence term by exploiting the property of the batch normalization.
Recall that batch normalization has been extensively used to make the training stable \citep{ioffe2015batch,hoffer2018norm,cho2017riemannian,miyato2018spectral,ulyanov2016instance}.
It has been consistently shown that the batch normalization is closely related to the norm of the Jacobian
matrix $\partial \Fb(\x) / \partial \x$, which is equal to $ \tilde \B(\x)\B(\x)^\top$ in our E-D CNNs.
For example, in their original paper \citep{ioffe2015batch}, the authors conjectured that ``Batch Normalization may
lead the layer Jacobians to have singular values close to 1,
which is known to be beneficial for training''.
By extending the idea in \citep{ioffe2015batch} to multiple layers, the batch normalization can be understood as to make the covariance
of the network output and input similar. For example,  for the uncorrelated input with
$\mathrm{Cov}[\x^{(i)}]=\sigma^2\Ib$, the batch normalization works to provide $\mathrm{Cov}[\Fb(\x^{(i)})]\simeq \sigma^2\Ib$.
Furthermore, for sufficiently smaller $\sigma$, we have
\begin{eqnarray*}
\sigma^2\Ib \simeq  \mathrm{Cov}[\Fb(\x^{(i)})]
 &=& \mathrm{Cov}\left[\tilde \B(\x^{(i)})\B(\x^{(i)})^\top\x^{(i)}\x^{\top(i)} \B(\x^{(i)})\tilde\B(\x^{(i)})^\top\right]\\
&=& \tilde \B(\x^{(i)})\B(\x^{(i)})^\top  \mathrm{Cov}\left[\x^{(i)}\x^{\top(i)}\right]\B(\x^{(i)})\tilde\B(\x^{(i)})^\top\\
&=& \sigma^2\tilde \B(\x^{(i)})\B(\x^{(i)})^\top \B(\x^{(i)})\tilde\B(\x^{(i)})^\top
\end{eqnarray*}
where the second equality comes  that within the small perturbation of the input, the corresponding frame representation
does not change \citep{ye2019cnn}.
Therefore, we have
\[\tilde \B(\x^{(i)})\B(\x^{(i)})^\top \simeq \Ib, \]
since $\tilde \B(\x^{(i)})\B(\x^{(i)})^\top $ is a square matrix. 
This suggests that
\begin{eqnarray*}
  \mathrm{div} E_L\{ \Fb(\x^{(i)},L) \} =    p\sum_i  \langle {\blmath b}_i(\x^{(i)}), \tilde  {\blmath b}_i(\x^{(i)}) \rangle=  p\mathrm{Tr}\left( \tilde \B(\x^{(i)})\B(\x^{(i)})^\top\right)    \simeq pn
\end{eqnarray*}
Since the resulting divergence term is just a constant,  the contribution of the divergence term in \eqref{eq:N2B} is considered trivial and can be neglected.

\subsection{Approximation of mean aggregation using an attention network}

Another important complication in calculating \eqref{eq:N2B} is that  the mean aggregation
$\widehat\mub(\x)=E_L\{\Fb(\x,L)\}$ is not available and we only have its empirical estimate.
Although the simplest way to obtain an empirical estimate is to average the overall results of the encoder-decoder CNNs, this may not be the best method because it does not reflect the data distribution of the results. Therefore, we propose a data attention network that efficiently combines all data so that
it  can adaptively incorporate neural network output from  various random sampling patterns.
More specifically, we use the following weighted average
\[\widetilde\mub_K(\x)= \sum_{k=1}^K w_k \Fb_\Thetab(\x, L_k),\quad \]
where $\{L_k\}_{k=1}^K$ denotes the bootstrap subsampling patterns and
$\{w_k\}_{k=1}^K$ is the corresponding weights.
To calculate the weight, we propose to use
the attention network illustrated in Fig. \ref{fig:flowchart}. 
Specifically, the input of the attention network is the $K$-dimensional
vector whose values are calculated by  an average pooling of $\Fb(\x,L_k)$.
This input is fed into a multilayer perceptron to generate the weight 
$w_k =w_k(\Xib)$, where $\Xib$ denotes the attention network parameters.

\subsection{Implementation Details}

\begin{figure}[!b]
  \center{
   \includegraphics[width=\textwidth]{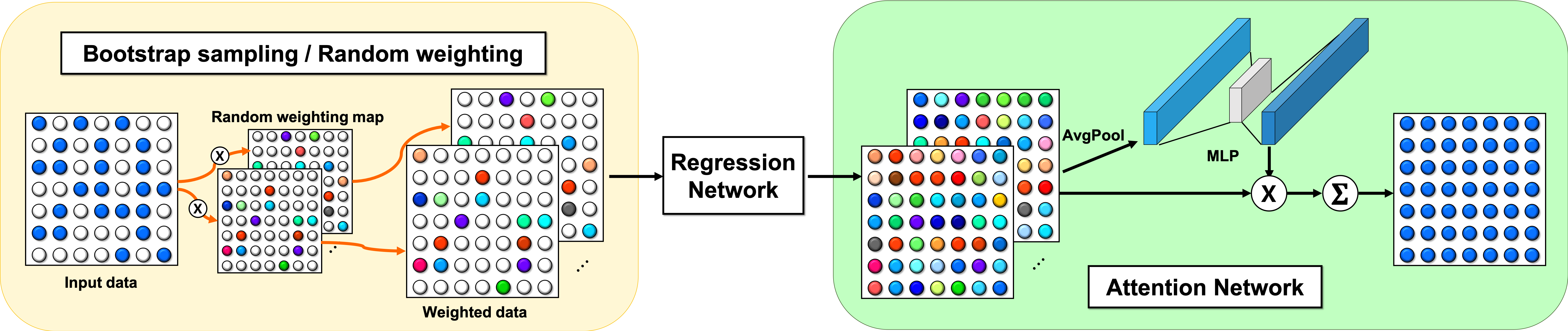}
}
  \caption{
  Concept of Noise2Boosting.
  The acquired data is boosted by bootstrap subsampled or multiplied with random weights. 
  The data can be restored by regression network and
  are aggregated by multiplying weight from the attention network.
  }
  \label{fig:flowchart}
\end{figure}

 A schematic diagram of the proposed method is illustrated in Figs. \ref{fig:flowchart}(a)(b). Our Noise2Boosting method  consists of three building blocks: 
bootstrap subsampling  or  random weighting, a regression network using encoder-decoder CNN, and an attention network. 

To estimate the weight $w_k$, our attention network consists of  two fully connected layer. The input dimension of the attention network is $\mathbb{R}^{1 \times 1 \times K}$ followed by the average pooling of the concatenated output of regression network. The number of hidden node is 64, and the final dimension of the output is $\mathbb{R}^{1 \times 1 \times K}$ for aggregation.


The  training was performed in two ways. 
  First, the neural network weight $\Thetab$ are first learned by minimizing the loss  \eqref{eq:N2V}.
 Then, the weight of $\Fb_\Theta(\x, L_k)$ is fixed and the attention network is trained by minimizing the loss with
 respect to  $\Xib$.
Second, both attention network and the E-D Networks are trained simultaneously by minimizing the loss:
\begin{eqnarray}\label{eq:N2Ball}
\ell_{N2B}\left(\Thetab,\Xib\right):=
\frac{1}{P}\sum_{i=1}^P \|\x^{(i)} - \sum_{k=1}^K w_k(\Xib) \Fb_\Thetab(\x, L_k)\|^2 
\end{eqnarray}
While  the first approach can reduce the computational
time and memory requirement,  this is only an approximation and  the second training scheme provides significantly better results.

The overall network was trained using Adam optimization \citep{kingma2014adam} with the momentum $\beta _1 = 0.9$ and $\beta_2 = 0.999$. The proposed network was implemented in Python using TensorFlow library \citep{abadi2016tensorflow} and trained using an NVidia GeForce GTX 1080-Ti graphics processing unit.

\section{Experimental Results}

Experiments were conducted for various inverse problems such as
compressed sensing MRI \citep{lustig2007sparse},  energy-dispersive X-ray spectroscopy (EDX) \citep{sole2007multiplatform} denoising,
and super-resolution.

U-net \citep{ronneberger2015u} was used as our regression network for compressed sensing MRI, and EDX denoising.
The network was composed of four stage with convolution, batch normalization, ReLU, and skip connection with concatenation. Each stage is composed of three $3 \times 3$ convolution layers followed by batch normalization and ReLU, except for the last layer, which is $1 \times 1 $ convolution layer. The number of convolutional filters increases from 64 in the first stage to 1024 in the final stage. 
For the case of super-resolution, we  employed Deep Back-Projection Network (DBPN) \citep{haris2018deep} which enables to restore the details by exploiting iteratively up- and down- sampling layer as the base algorithm for super-resolution task.

\begin{figure}[h]
  \center{
   \includegraphics[width=\textwidth]{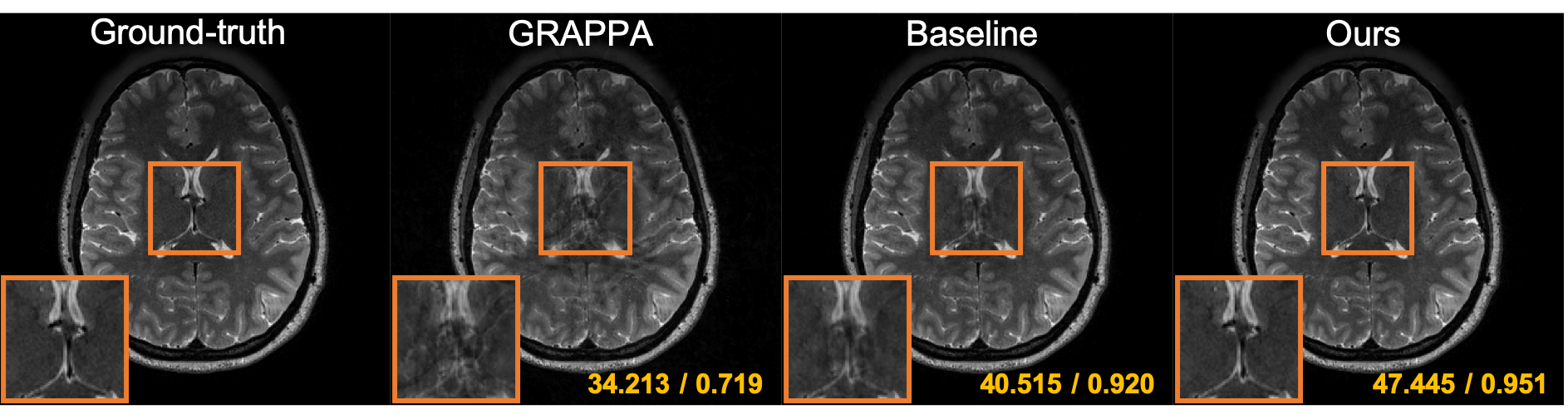}
}\vspace*{-0.5cm}
  \caption{Reconstruction results using MR dataset at the acceleration factor of $R = 13.45$. The PSNR and SSIM index values for each images are written at the corner. }
  \label{fig:mri_result}
\end{figure}
\begin{figure}[!h]
\vspace*{-0.5cm}
  \center{
   \includegraphics[width=\textwidth]{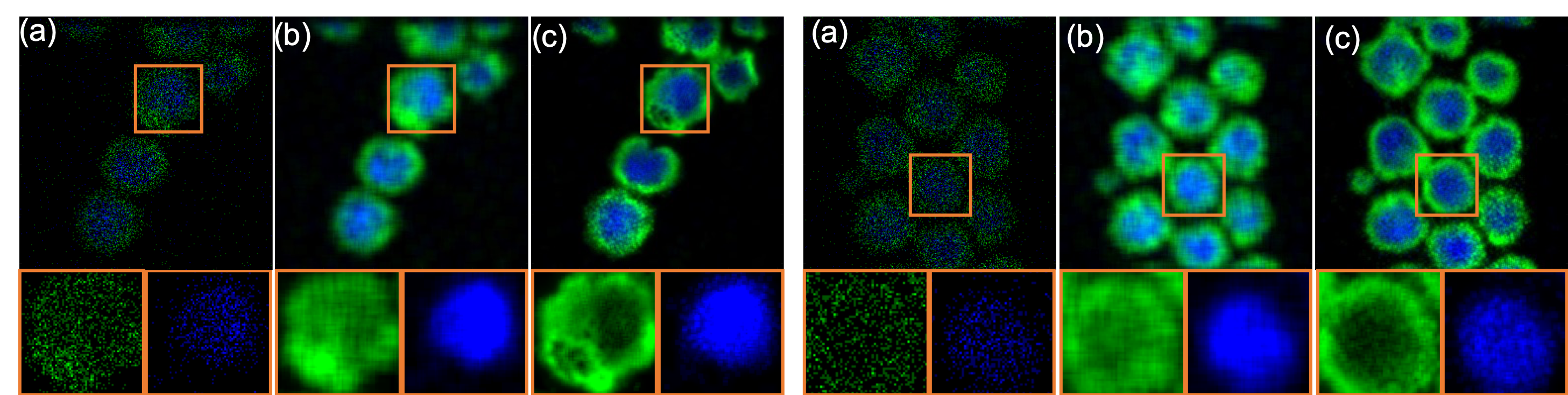}
}\vspace*{-0.5cm}
  \caption{Denoising EDX dataset. Green and blue particles refer to Zn and Cd, respectively. (a) Input data, (b) standard kernel-based denoising, and (c) Noise2Boosting.}
  \label{fig:edx_result}
\end{figure}
\begin{figure}[!h]
\vspace*{-0.5cm}
  \center{
   \includegraphics[width=\textwidth]{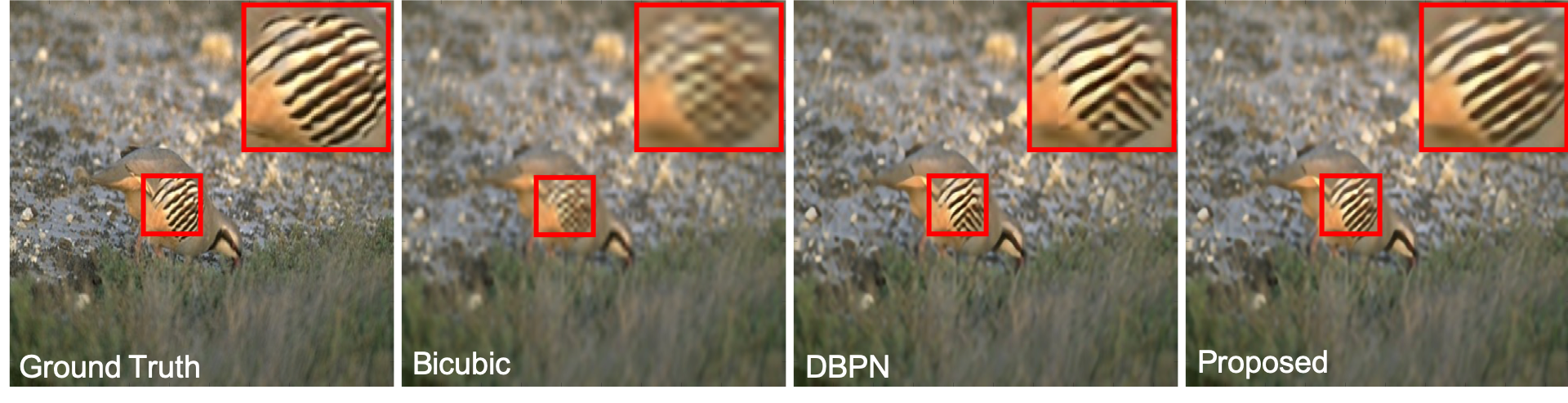}
}\vspace*{-0.5cm}
  \caption{Comparisons between (a) DBPN  and (b) Noise2Boosting for $\times4$ super-resolution task.}
  \label{fig:sr_result}
\end{figure}



In compressed sensing MRI \citep{lustig2007sparse}, the goal is to recover high quality MR images from sparsely sampled $k$-space data to reduce the acquisition time.  We performed supervised learning experiments with synthetic $k$-space data from Human Connectome Project (HCP) MR dataset (https://db.humanconnectome.org). Among the 34 subject data sets, 28 subject data sets were used for training and validation. The other subject data sets were used for test. 
As for the input for the neural network, we downsampled $k$-space with uniform sampling pattern, which corresponds to the acceleration factors of $R = 13.45$. The label images for neural network is the fully acquired $k$-space data.
A recent $k$-space learning algorithm \citep{han2018k} is employed as the  base deep learning algorithm which interpolates the missing elements in the $k$-space. We trained the baseline algorithm and the propose methods under the same conditions except for boostrap subsampling for $k$-space data in the proposed method. 
The number of random subsampling mask $K$ was set to 10, and the overall network are trained simultaneously by minimize the loss \eqref{eq:N2Ball}.
We also provide reconstruction results by GRAPPA \citep{griswold2002generalized}, which is a standard  $k$-space interpolation method in MRI.
We used six subjects of MR datasets to confirm the effectiveness of the proposed  method. 
Thanks to the boosting, the proposed method provided nearly perfect reconstruction results compared to other algorithms as shown in Fig. \ref{fig:mri_result}. 
 Moreover, it average peak signal-to-noise ratio (PSNR) and structural similarity (SSIM) index significantly outperform others 
 (PSNR/SSIM:  GRAPPA=34.52/0.72, Baseline =38.67dB/0.89, Noise2Boosting=\textbf{42.72dB/0.92}).

As for another experiment, we use the EDX data set which is mapped by STEM-EDX mode in transmission electron microscopy (TEM).
EDX is widely used for nano-scale quantitative and qualitative elemental composition analysis by measuring
x-ray radiation from the interaction with high energy electron and the material  \citep{mcdowell2012studying}.
However, the specimens can be quickly damaged by the high energy electrons, so the acquisition time should be  reduced to the minimum.
This usually results in very noisy and  even incomplete images as shown in Fig.~\ref{fig:edx_result}(a) and the goal is to denoise and interpolate the missing data. The main technical
difficulty is that there is no label data, so we need an unsupervised learning technique.
A widely used approach for EDX analysis is to use explicit average kernel as shown in Fig. \ref{fig:edx_result}(c). Unfortunately, this often results
in severe blurring when the measurement data is not sufficient.
Noise2Noise and Noise2Void do not work, since there are no specific noise models for the EDX.
In fact, this difficulty of EDX denoising was our original motivation for this work.
As for the input for training and inference, we use bootstrap sampled images from the measurement image in Fig.~\ref{fig:edx_result}(a) , and the network output is the measurement data
in Fig.~\ref{fig:edx_result}(a). 
We used 28 cases from the EDX dataset. 
The specimen are composed of quantum-dots,  where core and shell consist of Cadmium (Cd), Selenium (Se), Zinc (Zn), and Sulfur (S), respectively. 
For the bootstrap subsampling, the number of random subsampling mask was $K=30$. The regression network was optimized to minimize the loss \eqref{eq:N2V} with respect to $\Thetab$ first, after which the attention network was trained to properly aggregate the entire interpolated output. 
Since the regression network learn the measurement statistics, the network output provides more samples than the measured data and the attention network produces the  aggregated image. This produces sharper and accurate images as shown  in Fig. \ref{fig:edx_result}(c).

For the case of super-resolution,  the baseline network is trained using DIV2K \citep{agustsson2017ntire}, with totally 800 training images, on the $\times2$ and $\times4$ (in both horizontal and vertical directions) super-resolution, and training conditions were followed as described in \citep{haris2018deep}. For our Noise2Boosting training, the number of random subsampling mask $K$ was set to 32 and 8 for $\times2$ and $\times4$ task, respectively. In addition, the entire networks were trained simultaneously to minimize the loss \eqref{eq:N2Ball}. As described in Table \ref{table:sr_result}, our N2B method can improve the performance of the super-resolution task. Thanks to the bootstrapping and aggregation using attention network, the data distribution can be fully exploited to restore the high resolution components, which results in the properly reconstructed details of the image  as shown in Fig. \ref{fig:sr_result}.

\begin{table}[h]
  \caption{Comparison of PSNR and SSIM index for super-resolution task }
  \label{table:sr_result}
  \centerline{
   \resizebox{0.5\textwidth}{!}{ 
  \begin{tabular}{cccccc}
    \toprule
   & & \multicolumn{2}{c}{set14}  & \multicolumn{2}{c}{bsd100}                  \\
        \midrule
    Algorithm & Scale & PSNR & SSIM &  PSNR & SSIM \\
    \midrule
    DBPN \citep{haris2018deep} & 2 & 30.748 & 0.937 & 31.689 & 0.950 \\
       Proposed & 2 & \textbf{30.802} & \textbf{0.939} & \textbf{31.720} & \textbf{0.951}  \\    \bottomrule
  \end{tabular}
  }
     \resizebox{0.5\textwidth}{!}{ 
  \begin{tabular}{cccccc}
    \toprule
   & & \multicolumn{2}{c}{set14}  & \multicolumn{2}{c}{bsd100}                  \\
        \midrule
    Algorithm & Scale & PSNR & SSIM &  PSNR & SSIM \\
    \midrule
       DBPN \citep{haris2018deep} & 4 & 26.191 & \textbf{0.837 }& 26.803 & \textbf{0.855} \\
   Proposed & 4 & \textbf{26.222} & \textbf{0.837} & \textbf{26.855} & \textbf{0.855} \\    \bottomrule
  \end{tabular}
  }}
\end{table}

\section{Conclusion}

In this paper, we proposed a novel boosting scheme of  neural networks for various inverse problems with and without label data. 
Here, multiple input data were generated by bootstrap subsampling or random weight multiplication, after which final result are obtained
by aggregating the entire output of network using an attention network. 
Experimental results with
compressed sensing MRI,  electron microscopy image denoising and super-resolution showed
that the proposed method provides consistent improvement  for various inverse problems. 

\appendix

\section{Explicit Form of Encoder and Decoder Matrices}

The derivation in this paper is just a brief summary of \citep{ye2019cnn}, but included for self-containment.

Consider a symmetric E-D CNN  without skipped connection.
The encoder network maps a given input signal $\x\in\Xbc\subset \Rd^{d_0}$ to a 
 feature space $\z \in \Zbc\subset \Rd^{d_\kappa}$, whereas the decoder
 takes this feature map  as an input, process it  and produce an output 
$\y \in \Ybc\subset \Rd^{d_0}$.
At the $l$-th layer, $m_l$, $q_l$ and $d_l:=m_lq_l$ denote the dimension of the signal,  the number of filter channel, and the
total feature vector dimension, respectively.
%
%
%
Here,
the $j$-th channel output from the 
the $l$-th layer encoder can be represented by a multi-channel convolution operation:
 \begin{eqnarray}\label{eq:encConv}
\x_j^l = \sigma\left(\Phib^{l\top} \sum_{k=1}^{q_{l-1}}\left(\x_k^{l-1}\circledast \overline \psib_{j,k}^l\right)\right) ,
\end{eqnarray}
where  $\x_k^{l-1}$ refers to the $k$-th input channel signal, 
$\overline\psib_{j,k}^l\in \Rd^r$ denotes the $r$-tap convolutional kernel that is convolved with the $k$-th input channel to contribute
  to the  $j$-th channel output, and  $\Phib^{l\top}$ is the pooling operator.
Here, $\overline {\blmath v}$ is  the flipped version of the vector $\blmath v $ such that
$\overline v[n]= v[-n]$ with the periodic boundary condition, and $\circledast$ is the periodic convolution. 
The use of  the periodic boundary conditions
is to simplify the mathematical treatment of the boundary condition. 
Similarly,
 the $j$-th channel decoder layer convolution output is given by:
  \begin{eqnarray}\label{eq:decConv}
\tilde\x_j^{l-1} = \sigma\left(\sum_{k=1}^{q_{l}}\left(\tilde\Phib^l\tilde\x^{l}_k\circledast  {\tilde\psib_{j,k}^l}\right)\right) , 
\end{eqnarray}
where  $\tilde\Phib^l$ denotes the unpooling layer.
By concatenating the multi-channel signal in column direction as
 $$ \x^l:=\begin{bmatrix} \x^{l\top}_1 & \cdots & \x^{l\top}_{q_{l}} \end{bmatrix}^\top$$
the encoder and decoder convolution in  \eqref{eq:encConv} and \eqref{eq:decConv} can be represented using 
matrix notation:
\begin{eqnarray}\label{eq:ED}
 \x^l=\sigma(\Eb^{l\top} \x^{l-1})&,& \tilde \x^{l-1}=\sigma(\Db^l \tilde\x^{l})
\end{eqnarray}
where 
   \begin{eqnarray}\label{eq:El}
\E^l= \begin{bmatrix} 
\Phib^l\circledast \psib^l_{1,1} & \cdots &  \Phib^l\circledast \psib^l_{q_l,1}  \\
  \vdots & \ddots & \vdots \\
\Phib^l\circledast \psib^l_{1,q_{l-1}} & \cdots &  \Phib^l\circledast \psib^l_{q_{l},q_{l-1}}
 \end{bmatrix}
 \end{eqnarray}
 \begin{eqnarray}\label{eq:Dl}
 \Db^l= \begin{bmatrix} 
\tilde\Phib^l\circledast \tilde\psib^l_{1,1} & \cdots &  \tilde\Phib^l\circledast \tilde\psib^l_{1,q_l}  \\
  \vdots & \ddots & \vdots \\
\tilde\Phib^l\circledast \tilde\psib^l_{q_{l-1},1} & \cdots &  \tilde\Phib^l\circledast \tilde\psib^l_{q_{l-1},q_{l}}
 \end{bmatrix}
 \end{eqnarray}
and
$$\Phib^l=\begin{bmatrix}  \phib^l_1  & \cdots &  \phib^l_{m_l} \end{bmatrix},$$
\begin{eqnarray*}
\begin{bmatrix} \Phib^l \circledast \psib_{i,j}^l  \end{bmatrix}  :=\begin{bmatrix} \phib^l_1 \circledast \psib_{i,j}^l & \cdots & \phib^l_{m_l} \circledast  \psib_{i,j}^l\end{bmatrix}  \label{eq:defconv}
\end{eqnarray*}

\section{Proof of Propositions and Lemma}

\subsection{Proof of Proposition~1}

\begin{proof}
For a fixed sampling pattern $L$, $\Fb(\x,L)$ can be considered
as another neural network $\Gb(\x)$ with the sampling mask at the first layer.  Thus, we can easily see
$$E_{\x,\x^*}\{\|\x^*-\Fb(\x,L) \|^2|L\} =E_{\x,\x^*}\{\widehat{\mathrm{SURE}}(\x,L)|L\}.$$
Now, by taking expectation with respect to $L$, we conclude the proof.
\end{proof}

\subsection{Proof of Proposition~2}

\begin{proof}
Note that
\begin{eqnarray*}
 E_{L}\{\|\x^* - \Fb(\x,L) \|^2 |\x,\x^*\}&=&  \x^{*\top}\x^* - 2\x^{*\top}  E_{L} \{\Fb(\x,L) \}+ E_{L}\{ \|\Fb(\x,L) \|^2\} \\
 &\geq&  \x^{*\top}\x^* - 2\x^{*\top}    E_{L}\{ \Fb(\x,L) \}+ \|E_{L} \{\Fb(\x,L)  \}\|^2\\
 &=& \left\|\x^* - E_{L}\{ \Fb(\x,L) \}\right\|^2
\end{eqnarray*}
where we use the Jensen's inquality for the inequality.
Then, we have
\begin{eqnarray*}
 E_{\x,\x^*,L}\|\x^* - \Fb(\x,L) \|^2 & = &E_{\x,\x^*} E_{L}\{\|\x^* - \Fb(\x,L) \|^2 |\x,\x^*\}\\
 & \geq& E_{\x,\x^*} \left\|\x^* - E_{L}\{ \Fb(\x,L) \}\right\|^2
\end{eqnarray*}
\end{proof}

\subsection{Proof of Lemma~1}
\begin{proof}
In Proposition 6 of \citep{ye2019cnn},  it was shown that
$\partial \Fb(\x) / \partial \x = \tilde\B(\x)\B(\x)^\top$ for the case of ReLU network.
Accordingly, we have
\begin{eqnarray*}
 \mathrm{div} \{ \Fb(\x) \}  =  \mathrm{Tr}\left( \frac{\partial \Fb(\x)}{\partial \x} \right) & =&  \mathrm{Tr}\left(\tilde \B(\x)\B(\x)^\top\right) 
=  \mathrm{Tr}\left(\B(\x)^\top\tilde \B(\x)\right)  \\
&=&  \sum_i  \langle {\blmath b}_i(\x), \tilde  {\blmath b}_i(\x) \rangle
\end{eqnarray*}
where $\mathrm{Tr}(\A)$ denotes the trace of $\A$. 
This concludes the proof.
\end{proof}

\subsection{Proof of Proposition~3}

\begin{proof}
Note that subsampling  is equivalent to multiply $\{0,1\}$ mask to the input vector.
Therefore, the corresponding neural network output can be represented by
\begin{eqnarray}\label{eq:basisL}
\y := \Fb(\x,L) 
= \sum_{i} \langle {\blmath b}_i(\x), \Lb \x \rangle \tilde  {\blmath b}_i(\x) = \tilde \B(\x)\B(\x)^\top \Lb \x
\end{eqnarray}
Here, $\Lb$ denotes a  diagonal matrix  with $\{0,1\}$ values, where the probability of 1 is  $p$.
Accordingly, we have
\begin{eqnarray*}
 \mathrm{div} E_L\{ \Fb(\x,L) \}   &= &  \mathrm{div}\{ \tilde \B(\x)\B(\x)^\top E_L\{\Lb \x\}\}  \\
 &=& p  \mathrm{div} \{ \Fb(\x) \} =  p\sum_i  \langle {\blmath b}_i(\x), \tilde  {\blmath b}_i(\x) \rangle
\end{eqnarray*}
where we use $E_L\{\Lb\x\}= p\x$. The proof for the boosted samples is basically the same. This concludes the proof.
\end{proof}

\section{Additional Results}

Recall that one of the main limitations of Noise2Noise \citep{lehtinen2018noise2noise} (N2N) is that
it requires many noisy realizations for the same images, which is not usually feasible in many applications.
Otherwise, during the training of N2N, different noises are separately added for each epoch to the clean image to generates multiple input and label with different noise
realizations. 
This implies that the ground-truth clean images are necessary for N2N training unless multiple noisy measurements are available.

On the other hand, 
%
%
we only require {\em single} realization of noisy images for the training of our Noise2Boosting (N2B). 
The main idea is that,  as discussed before,  pixel-wise random weight was multiplied to the noisy input data for boosting. The value of random weight is randomly chosen between $0.8$ and $1.2$.
In order to diversify the image data, we found that the noisy augmentation was helpful. 
Specifically, we randomly generated  Gaussian noises with noise standard deviation $\sigma \in [10, 40]$ 
and add them to the noisy measurement data, before applying the random weighting.
Note that this procedure is fundamentally different from Noise2Noise which adds noises to the {\em clean} images, since our augmentation
add the noisy data to the {\em noisy} measurement. Therefore, we do not need either many noisy realization of the same clean image nor
the clean image to synthetically generate the noisy output.

For a fair comparison,  a standard U-net was employed for N2N and N2B networks using DIV2K \citep{agustsson2017ntire}, with totally 800 training images.  The initial learning rate for both networks was set to 0.0003, and divided by half per 50 epochs until it reached approximately 0.00001. Minibatch size of 8 was used in all experiments. 
As explained in \citep{lehtinen2018noise2noise}, the synthetic Gaussian noise was added to the clean image for training N2N. 
Accordingly, during the training of N2N,  the standard deviation of addictive Gaussian noise is randomly selected between 0 and 50. Furthermore, the different noise are separately added to clean image as input and label for training network, respectively. 

\begin{figure}[h]
  \center{
   \includegraphics[width=\textwidth]{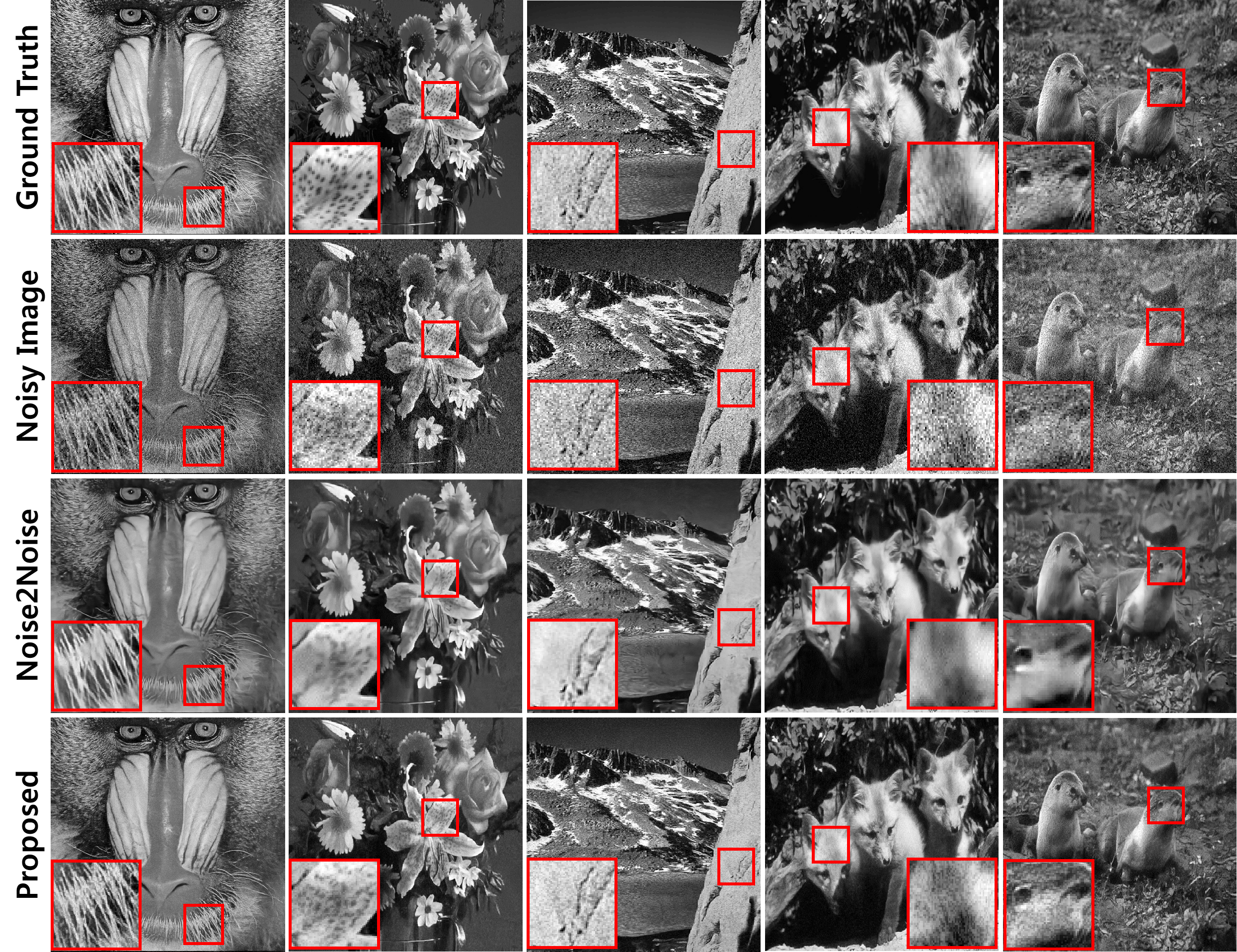}
}
  \caption{
Comparisons between Noise2Noise and Noise2Boosting for Gaussian denoising task with $\sigma = 25$.
  }
  \label{fig:denoise_result}
\end{figure}

For inference, we used the single noise image with $\sigma=25$ for N2N.  For N2B, we use $K=2$ boosted samples.
As shown in Table \ref{table:denoising_result}, although our N2B network does not need clean image for training, the proposed N2B network generally outperformed N2N network.  In addition, the proposed N2B network can restore better details and texture than N2N network as shown in Fig. \ref{fig:denoise_result} .

\begin{table}[h]
  \caption{Comparison of PSNR and SSIM index for Gaussian noise ($\sigma$ = 25)}
  \label{table:denoising_result}
  \centering
  \begin{tabular}{ccccccc}
    \toprule
 & \multicolumn{2}{c}{Noisy Image}  & \multicolumn{2}{c}{Noise2Noise \citep{lehtinen2018noise2noise} }  & \multicolumn{2}{c}{Proposed}    \\
    \midrule
   & PSNR & SSIM &  PSNR & SSIM &  PSNR & SSIM \\
   \midrule
set5 & 20.620 & 0.353 & \textbf{29.180} & \textbf{0.756} & 29.117 & 0.742 \\
set14 & 20.527 & 0.395 & 27.847 & 0.711 & \textbf{28.049} & \textbf{0.725} \\
bsd & 20.477 & 0.390 & 27.200 & 0.691 & \textbf{27.648} & \textbf{0.715} \\
    \bottomrule
  \end{tabular}
\end{table}

%
%


\end{document}